\documentclass[sigconf,anonymous=false,review=false]{acmart}

\usepackage[skip=0pt]{caption}
\usepackage[inline]{enumitem}
\usepackage{multirow}
\usepackage{acronym}
\acrodef{TDS}{Task-oriented Dialogue System}
\acrodef{MTDS}{Modular Task-oriented Dialogue System}
\acrodef{MOE}{Mixture-of-Expert}
\acrodef{NMOE}{Neural Mixture-of-Expert}
\acrodef{KB}{Knowledge Base}
\acrodef{SeqMoE}{Sequence-level Mixture-of-Expert}
\acrodef{TokenMoE}{Token-level Mixture-of-Expert}
\acrodef{SentenceMoE}{Sentence-level Mixture-of-Expert}
\acrodef{MLP}{Multilayer Perceptron}
\acrodef{RNN}{Recurrent Neural Network}
\acrodef{GRU}{Gated Recurrent Unit}
\acrodef{Seq2Seq}{Sequence-to-Sequence}
\acrodef{Seq2SeqAttn}{Sequence-to-Sequence with Attention}
\acrodef{LSTM}{Long Short-Term Memory}
\acrodef{S2SAttnLSTM}{Sequence-to-Sequence with Attention Using LSTM}
\acrodef{MultiWOZ}{Multi-Domain Wizard-of-Oz}

\acrodef{NLU}{Natural Language Understanding}
\acrodef{DST}{Dialogue State Tracking}
\acrodef{PL}{Policy Learning}
\acrodef{NLG}{Natural Language Generation}

\acrodef{POMDP}{Partially Observable Markov Decision Process}

\usepackage{amsmath}

\DeclareMathOperator{\softmax}{softmax}
\DeclareMathOperator{\MLP}{MLP}

\newcommand{\jpei}[1]{\textcolor{black}{#1}}

\AtBeginDocument{%
  \providecommand\BibTeX{{%
    \normalfont B\kern-0.5em{\scshape i\kern-0.25em b}\kern-0.8em\TeX}}}

\setcopyright{acmcopyright}
\copyrightyear{2019}
\acmYear{2019}
\acmDOI{xx.xxx/xxx_x}

\acmConference[WCIS '19]{WCIS '19: 1st Workshop on Conversational Interaction Systems}{July 25, 2019}{Paris, France}
\acmPrice{15.00}
\acmISBN{978-1-4503-9999-9/18/06}

\acmSubmissionID{5}

\begin{document}

\title[A Modular \acl{TDS}]{A Modular \acl{TDS} \\ Using a \aclp{NMOE}}

\author{Jiahuan Pei}
\affiliation{%
   \institution{University of Amsterdam}
   \city{Amsterdam}
   \country{The Netherlands}
}
\email{j.pei@uva.nl}

\author{Pengjie Ren}
\affiliation{%
   \institution{University of Amsterdam}
   \city{Amsterdam}
   \country{The Netherlands}
}
\email{p.ren@uva.nl}

\author{Maarten de Rijke}
\affiliation{%
   \institution{University of Amsterdam}
   \city{Amsterdam}
   \country{The Netherlands}
}
\email{derijke@uva.nl}

\begin{abstract}
End-to-end \acp{TDS} have attracted a lot of attention for their superiority (e.g., in terms of global optimization) over pipeline modularized \acp{TDS}.
Previous studies on end-to-end \acp{TDS} use a single-module model to generate responses for complex dialogue contexts.
However, no model consistently outperforms the others in all cases.

We propose a neural \ac{MTDS} framework, in which a few expert bots are combined to generate the response for a given dialogue context.
\ac{MTDS} consists of a chair bot and several expert bots.
Each expert bot is specialized for a particular situation, e.g., one domain, one type of action of a system, etc. 
The chair bot coordinates multiple expert bots and adaptively selects an expert bot to generate the appropriate response.
We further propose a \ac{TokenMoE} model to implement \ac{MTDS}, where the expert bots predict multiple tokens at each timestamp and the chair bot determines the final generated token by fully taking into consideration the outputs of all expert bots.
Both the chair bot and the expert bots are jointly trained in an end-to-end fashion.

To verify the effectiveness of \ac{TokenMoE}, we carry out extensive experiments on a benchmark dataset.
Compared with the baseline using a single-module model, our \ac{TokenMoE} improves the performance by 8.1\% of \textit{inform rate} and 0.8\% of \textit{success rate}.
\end{abstract}



\maketitle

\section{Introduction}

As an important branch of spoken dialogue systems, \acfp{TDS} have raised considerable interest due to their broad applicability, e.g., for booking flight tickets or scheduling meetings~\cite{young2013pomdp,williams2017hybrid}. 
Unlike open-ended dialogue systems~\cite{serban2016building}, \acp{TDS} aim to assist users to achieve specific goals.

Existing \ac{TDS} methods can be divided into two broad categories: modularized pipeline \acp{TDS}~\cite{end2end_dataset_paper_babi_bordes,chen2017survey,young2013pomdp} and end-to-end single-module \acp{TDS}~\cite{eric2017key,wen2016network}.
The former decomposes the task-oriented dialogue task into modularized pipelines that are addressed by separate models while the latter proposes to use an end-to-end model to solve the task.
End-to-end single-module \acp{TDS} have many attractive characteristics, e.g., global optimization and easier adaptation to new domains~\cite{chen2017survey}.
\jpei{However, existing studies on end-to-end single-module \acp{TDS} mostly generates a response token by token, where each token is drawn from only one distribution over output vocabulary. 
We think this is unreasonable because the distribution differs a lot among different intents. }
Actually, more and more empirical studies from different machine learning applications suggest that no model consistently outperforms all others in all cases~\cite{dietterich2000ensemble,masoudnia2014mixture}.

\begin{figure}[h]
    \vspace*{-0.8\baselineskip}
    \centering
    \begin{tabular}{c}
        \includegraphics[width=0.95\columnwidth]{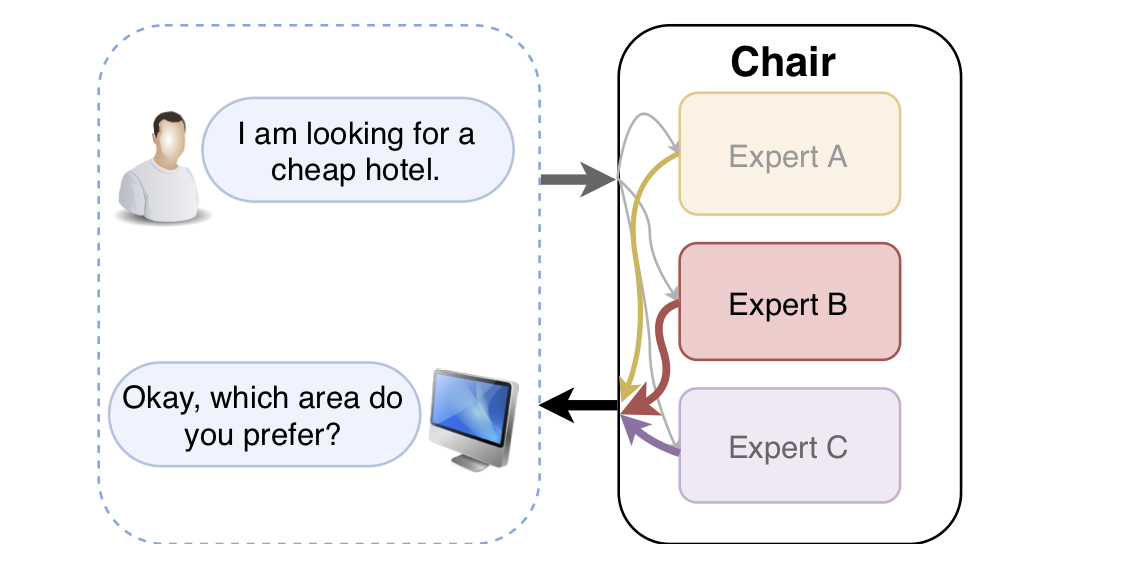}
    \end{tabular}
    \caption{\acfi{MTDS} framework.}
    \label{fig:frameworks}
        \label{fig:tds_pipeline}
        \label{fig:tds_end2end}
        \label{fig:tds_modular}
    \vspace*{-1\baselineskip}
\end{figure}

Inspired by this intuition, we propose a new \acfi{MTDS} framework, as shown in Fig.~\ref{fig:frameworks}.
\ac{MTDS} consists of a \emph{chair bot} and several \emph{expert bots}.
\begin{figure*}[t]
    \centering
    \includegraphics[width=0.85\textwidth]{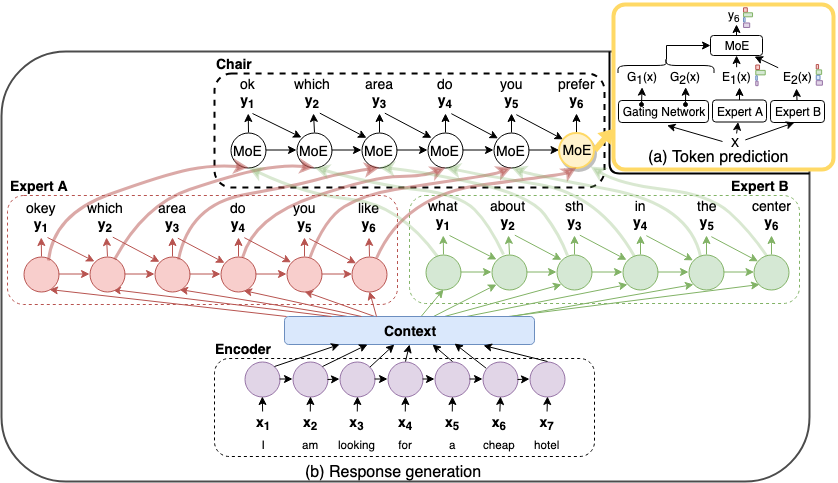}
    \caption{Overview of \ac{TokenMoE}. Figure (a) illustrates how does the model generate the token $y_6$ given sequence $X$ as an input. Figure (b) shows how does the model generate the whole sequence $Y$ as a dialogue response.}
    \label{fig:token_moe}
    \vspace*{-1\baselineskip}
\end{figure*}
Each expert bot is specialized for a particular situation, e.g., one domain, one type of action of a system, etc. 
The chair bot coordinates multiple expert bots and adaptively selects an expert bot to generate the final response.
Compared with existing end-to-end single-module \acp{TDS}, the advantages of \acp{MTDS} are two-fold. 
First, the specialization of different expert bots and the use of a dynamic chair bot for combining the outputs breaks the bottleneck of a single model.
Second, it is more easily \emph{traceable}: we can analyze who is to blame when the model makes a mistake.
Under this framework, we further propose a neural \acfi{MOE} model, namely \acfi{TokenMoE}, where the expert bots predict multiple tokens at each timestamp and the chair bot determines the final generated token by taking into account the outputs of all expert bots. 
We devise a global-and-local learning scheme to train \ac{TokenMoE}.
We design a \emph{localized expert loss} to force each expert to specialize on a particular task.
We also design a \emph{global chair loss} to differentiate the loss incurred from different experts~\cite{guo2018multi}.

To verify the effectiveness of \ac{TokenMoE}, we carry out extensive experiments on the MultiWOZ benchmark dataset.
The results show that, compared with the baseline using a single-module model, our \ac{TokenMoE} improves the performance by 8.1\% of \textit{inform rate}, and 0.8\% of \textit{success rate}. 

The contributions of this paper can be summarized as follows.
\begin{itemize}[leftmargin=*,align=left,nosep]
    \item We propose the \ac{MTDS} framework, which breaks the bottleneck of a single-module model and provides better traceability of mistakes.
    \item We present the \ac{TokenMoE} model to implement \ac{MTDS} at the token-level.
    \item We devise a global-and-local learning scheme to effectively train \ac{TokenMoE}.
\end{itemize}
\section{Methodology}
\label{sec:tmoe}
Let $\mathcal{D}=\{(X_p, Y_p)\}_{p=1}^{|\mathcal{D}|}$ denote a dataset with $|\mathcal{D}|$ independent random samples of $(X, Y)$, where $X = (x_1, \dots, x_m)$ is a sequence of dialogue context with $m$ words and $Y = (y_1, \dots, y_n)$ is a sequence of system response with $n$ words. 
The model aims to optimize the generation probability of $Y$ conditioned on $X$, i.e., $p(Y|X)$.

\vspace*{-0.5\baselineskip}
\subsection{$\ac{MTDS}$ framework}
We propose the \ac{MTDS} framework, which consists of two types of modules as shown in Figure~\ref{fig:frameworks}:
\begin{itemize}[leftmargin=*,align=left,nosep]
    \item $k$ \textbf{expert} bots, each of which is specialized for a particular situation, namely \textit{intent} (e.g., one domain, one type of action of a system, etc.).
    Those intents partition dataset $\mathcal{D}$ into $k$ pieces $\mathcal{S}=\{\mathcal{S}_l\}_{l=1}^{k}$, where $\mathcal{S}_l \triangleq \{(X^l, Y^l)\}$.
    Each expert is trained to predict $\mathbf{p}^l (Y^l|X^l)$.
    We expect the $l$-{th} expert generally performs better than the others on $\mathcal{S}_l$.

    \item a \textbf{chair} bot, which learns to coordinate a group of expert bots to make an optimal decision.
    The chair bot is trained to predict $\mathbf{p}(Y|X)$, where $(X, Y)$ is any sample pair from $\mathcal{D}$.
\end{itemize}

\vspace*{-0.5\baselineskip}
\subsection{$\acs{TokenMoE}$ model}
In this section, we introduce $\acs{TokenMoE}$, a token-level implementation of the $\ac{MTDS}$ framework.
As shown in Figure~\ref{fig:token_moe}, $\acs{TokenMoE}$ consists of three types of components, i.e., a shared encoder, $k$ expert decoders, and a chair decoder.

\subsubsection*{\textbf{Shared context encoder.}}
The role of shared context encoder is to read the dialogue context sequence and construct their representations at each timestamp.
Here we follow \citet{budzianowski2018multiwoz} and employ a \ac{LSTM} \cite{Hochreiter:1997:LSM:1246443.1246450} to map the input sequence $X$ to hidden vectors $\{\mathbf{h}_1, \dots, \mathbf{h}_m\}$.
The hidden vector $\mathbf{h}_i$ at timestep $i$-{th} can be represented as:
\begin{equation}
\mathbf{h}_i, \mathbf{s}_i =\text{LSTM}(emb(x_i), \mathbf{h}_{i-1}, \mathbf{s}_{i-1}),
\end{equation}
where $emb(x_i)$ is the embedding of the token $x_i$ at $i$.
The initial state of \ac{LSTM} $\mathbf{s}_{0}$ is set to 0.

\subsubsection*{\textbf{Expert decoder.}}
The $l$-{th} expert outputs the probability $\mathbf{p}_j^l$ over the vocabulary set $\mathcal{V}$ at $j$-{th} step by:
\begin{equation}\label{expertp}
\begin{split}
&\mathbf{p}_j^l = \softmax(\mathbf{U}^{T}\mathbf{o}_j^l+\mathbf{a}),\\
&\mathbf{o}_j^l, \mathbf{s}_j^l = \text{LSTM}(y_{j-1}^l \oplus \mathbf{c}^{l}_j, \mathbf{o}_{j-1}^l, \mathbf{s}_{j-1}^l),
\end{split}
\end{equation}
where $\mathbf{U}, \mathbf{a}$ are learnable matrices. $\mathbf{s}_j^l$ is the state vector which is initialized by the last state of the shared context encoder.
$y_{j-1}^l$ is the generated token at timestamp $j-1$ by expert $l$.
$\mathbf{c}^l_j$ is the context vector which is calculated with a concatenation attention mechanism \cite{bahdanau2014neural,D15-1166} over the hidden representations from shared context encoder.
\begin{equation}
\begin{split}
\mathbf{c}^{l}_j & = \sum_{i=1}^{m}\alpha^{l}_{ji}\mathbf{h}_i, \\
\alpha^{l}_{ji} & = \frac{\exp(w^{l}_{ji})}{\sum_{i=1}^{m} \exp({w^{l}_{ji})}},\\
w^{l}_{ji} & = \mathbf{v}^T\tanh{(\mathbf{W}^T(\mathbf{h}_i \oplus \mathbf{s}^{l}_{j-1})+\mathbf{b})},
\end{split}
\end{equation}
where $\alpha$ is the attention weights;
$\oplus$ is the concatenation operation.
$\mathbf{W}$, $\mathbf{b}$, $\mathbf{v}$ are learnable parameters, which are not shared by different experts in our experiments.

\subsubsection*{\textbf{Chair decoder.}}
The chair decoder estimates the final token prediction distribution $\mathbf{p}_j$ by combining prediction distribution of all experts (including chair itself) with a proposed token-level \ac{MOE} scheme.
As shown in Fig.~\ref{fig:token_moe}(b), following the typical neural \ac{MOE} architecture~\cite{shazeer2017outrageously,schwab2019granger}, $\mathbf{p}_j$ is computed based on the state $\mathbf{s}_j^l$ and token prediction distribution $\mathbf{p}_j^l$ from all experts (including chair itself) at $j$ as follows.
\begin{equation}\label{chairp}
\mathbf{p}_j = \sum_{l=1}^{k+1} \beta_{j}^{l} \cdot \mathbf{p}_j^l,
\end{equation}
where $\mathbf{p}_j^{k+1}$ is the prediction of the chair, which employs the same architecture of the other experts but is trained on all data.
$\beta_{j}^{l}$ is the normalized importance scores that can be computed as:
\begin{equation}
\label{beta}
\begin{split}
&\beta_{j}^{l} = \frac{\exp(u_l^T u_{e, l})}{\sum_{b=1}^{k} \exp(u_b^T u_{e, l})},\\
&u_l = \MLP (\mathbf{h}),\\
&\mathbf{h} = \mathbf{s}_j^1 \oplus \mathbf{p}_j^1 \oplus \dots \oplus \mathbf{s}_j^k \oplus \mathbf{p}_j^k \oplus \mathbf{s}_j^{k+1} \oplus \mathbf{p}_j^{k+1},
\end{split}
\end{equation}
where $u_{e,l}$ is an expert-specific, learnable vector that reflects which dimension of the projected hidden representation is highlighted for the expert. 

\subsubsection*{\textbf{Loss function.}}
We devise a global-and-local learning scheme to train \ac{TokenMoE}.
Each expert $l$ is optimized by a localized expert loss defined on $\mathcal{S}_l$, which forces each expert to specialize on one of the portions of data $\mathcal{S}_l$.
We use cross-entropy loss for each expert and the joint loss for all experts are as follows.
\begin{equation}\label{expertloss}
\mathcal{L}_\mathit{experts} = \sum_{l=1}^{k+1} \sum_{(X^l, Y^l) \in \mathcal{S}_l} \sum_{j=1}^{n} \mu_{k} y_j^l \log \mathbf{p}_j^l,
\end{equation}
where $p_j^{l}$ is the token prediction by expert $l$ (Eq.~\ref{expertp}) computed on the $r$-th data sample;
$y_j^{l, r}$ is a one-hot vector indicating the ground truth token at $j$; and
$\mu_{k}$ is the weight of the $k$-{th} expert.

We also design the global chair loss to differentiate the loss incurred from different experts. 
The chair can attribute the source of errors to the expert in charge.
For each data sample in $\mathcal{D}$, we follow the \ac{MOE} architecture and calculate the combined taken prediction $\mathbf{p}_j$ (Eq.~\ref{chairp}). 
Then the total loss incurred by \ac{MOE} can be denoted as follows.
\begin{equation}
\begin{split}
\mathcal{L}_\mathit{chair} 
&=\sum_{r=1}^{|\mathcal{D}|} \sum_{j=1}^{n} y_j \log \mathbf{p}_j.
\end{split}
\end{equation}
Our overall optimization follows the joint learning paradigm that is defined as a weighted combination of constituent loss.
\begin{equation}\label{finalloss}
\mathcal{L} = \lambda \cdot \mathcal{L}_\mathit{experts} + (1 - \lambda) \cdot \mathcal{L}_\mathit{chair},
\end{equation}
where $\lambda$ is a hyper-parameter.

\section{Experimental Setup}
\label{sec:experimental-setup}

\vspace*{-0.5\baselineskip}
\subsection{Research questions}
We seek to answer the following research questions.

\begin{enumerate}[label=(\textbf{RQ\arabic*}),leftmargin=*]
\item Is there a single model that consistently outperforms the others on all domains? The point of this question is to verify the motivation behind \ac{MTDS} and \ac{TokenMoE}. 

\item Does the \ac{TokenMoE} model outperform the state-of-the-art end-to-end single-module \ac{TDS} model? The point of this question is to determine the effectiveness of the proposed \ac{TokenMoE} model.

\item How do the proposed token-level \ac{MOE} scheme (Eq.~\ref{chairp} and Eq.~\ref{beta}) and the global-and-local learning scheme (Eq.~\ref{expertloss} and Eq.~\ref{finalloss}) in the \ac{TokenMoE} model affect the final performance? The point of this question is to do an ablation study on effective learning schemes.
\end{enumerate}

\vspace*{-0.5\baselineskip}
\subsection{Comparison methods}
\label{subsec:bsls}

\begin{table}[t]
\centering
\caption{Different settings of learning schemes.}
\label{tab:moe_variants}
\begin{tabular}{cccc}
\toprule
Variants & $\mathbf{p}_{j}$ (Eq.~\ref{expertp}) & $\mu_{k}$ (Eq.~\ref{expertloss})    & $\lambda$ (Eq.~\ref{finalloss}) \\ 
\midrule
S1       & \acs{MOE} & learnable      & learnable \\ 
S2       & \acs{MOE} & -- & 0.0       \\ 
S3       & w/o \acs{MOE}& $\frac{1}{k}$     & 0.5 \\ 
S4       & \acs{MOE} & $\frac{1}{k}$ & 0.5 \\
\bottomrule 
\end{tabular}
\end{table}

\begin{table*}[]
\setlength{\tabcolsep}{0.8pt}
\caption{Performance of the single-module baseline (\acs{S2SAttnLSTM}) and its three variations (V1, V2 and V3) on different domains. Bold highlighted results indicate a statistically significant improvement of a metric over the strongest baseline on the same domain (paired t-test, $p < 0.01$). UNK denotes a unknown domain excluding the domains described in \S\ref{subsec:dataset}. Please note that the number of the evaluated dialogue turns varies among different domains.}
\label{tab:bsls_on_domain}
\centering
\begin{tabular}{lrrrrrrrrrrrrrrrrc}
\toprule
           & \multicolumn{4}{c}{\textbf{Inform (\%)}} & \multicolumn{4}{c}{\textbf{Success (\%)}} & \multicolumn{4}{c}{\textbf{BLEU (\%)}}   & \multicolumn{4}{c}{\textbf{Score}}  & \multirow{2}{*}{\begin{tabular}[r]{@{}l@{}}\# of  turns\end{tabular}} \\
           & Baseline   & /V1      & /V2      & /V3      & Baseline   & /V1       & /V2      & V3      & Baseline & /V1     & /V2     & /V3     & Baseline & /V1     & /V2     & /V3     &                                                                                            \\
\midrule
Attraction & 87.20      & 86.20   & \textcolor{red}{\textbf{91.80}}   & 88.70   & 81.30      & 74.80    & \textcolor{red}{\textbf{83.70}}   & 83.70   & 15.14   & 14.95 & \textcolor{red}{\textbf{16.08}} & 14.86 & 99.39    & 95.45  & \textcolor{red}{\textbf{103.83}} & 101.06 & 1042                                                                                       \\
Hotel      & 89.90      & \textcolor{blue}{\textbf{93.90}}   & 89.90   & 90.30   & 87.50      & \textcolor{blue}{\textbf{91.70}}    & 87.40   & 89.10   & \textcolor{blue}{\textbf{16.60}}   & 15.60 & 15.11 & 14.13 & 105.30   & \textcolor{blue}{\textbf{108.40}} & 103.76 & 103.83 & 1068                                                                                       \\
Restaurant & 89.20      & \textcolor{magenta}{\textbf{91.70}}   & 86.40   & 86.10   & 85.80      & \textcolor{magenta}{\textbf{87.80}}    & 84.00   & 83.40   & 17.07   & \textcolor{magenta}{\textbf{17.70}} & 16.07 & 17.34 & 104.57   & \textcolor{magenta}{\textbf{107.45}} & 101.27 & 102.09 & 1024                                                                                       \\
Taxi       & 100.00     & 100.00  & 100.00  & 100.00  & \textcolor{cyan}{\textbf{99.90}}     & 99.80    & \textcolor{cyan}{\textbf{99.90}}   & 99.80   & 17.33   & 19.18 & \textcolor{cyan}{\textbf{20.13}} & 18.32 & 117.28   & 119.08 & \textcolor{cyan}{\textbf{120.08}} & 118.22 & 395                                                                                        \\
Train      & 77.70      & 77.70   & 79.00   & \textcolor{orange}{\textbf{81.60}}   & 75.60      & 74.80    & 77.20   & \textcolor{orange}{\textbf{79.60}}   & 20.35   & 15.64 & \textcolor{orange}{\textbf{22.81}} & 20.62 & 97.00    & 91.89  & 100.91 & \textcolor{orange}{\textbf{101.22}} & 1702                                                                                       \\
Booking    & 100.00     & 100.00  & 100.00  & 100.00  & 100.00     & 100.00   & 100.00  & 100.00  & 22.05   & 21.61 & 21.96 & \textcolor{teal}{\textbf{22.06}} & 122.05   & 121.61 & 121.96 & \textcolor{teal}{\textbf{122.06}} & 1407                                                                                       \\
General    & 100.00     & 100.00  & 100.00  & 100.00  & 100.00     & 100.00   & 100.00  & 100.00  & 20.21   & 19.53 & 20.13 & \textcolor{violet}{\textbf{20.80}} & 120.21   & 119.53 & 120.13 & \textcolor{violet}{\textbf{120.80}} & 2596                                                                                       \\
UNK        & 100.00     & 100.00  & 100.00  & 100.00  & 100.00     & 100.00   & 100.00  & 100.00  & 12.40   & 11.75 & \textcolor{purple}{\textbf{13.12}} & 11.80 & 112.40   & 111.75 & \textcolor{purple}{\textbf{113.12}} & 111.80 & 81  \\                                                                                      
\bottomrule
\end{tabular}
\vspace*{-0.8\baselineskip}
\end{table*}


We use the dominant \ac{Seq2Seq} model in an encoder-decoder architecture~\cite{chen2017survey} and reproduce the state-of-the-art single model baseline, namely \ac{S2SAttnLSTM}~\cite{budzianowski2018multiwoz,budzianowski2018towards}, based on the source code provided by the authors.\footnote{\url{https://github.com/budzianowski/multiwoz}. For fair comparison, we remove validation set from training set and report the reproduced results.}

To answer \textbf{RQ1}, we investigate the performance of the following variants of \ac{S2SAttnLSTM} on different domains. 

\begin{itemize}[leftmargin=*,align=left,nosep]
\item \textbf{V1.} This variant excludes the attention mechanism from the baseline model and keeps the other settings unchanged.
\item \textbf{V2.} This variant changes the LSTM cell as GRU and keeps the other settings the same.
\item \textbf{V3.} This variant reduces the number of hidden units to 100 and maintains the other settings.
\end{itemize}
 
\noindent%
To answer \textbf{RQ2}, we train \ac{TokenMoE} based on the benchmark dataset and test how it performs compared to the single-module baseline.

To answer \textbf{RQ3}, we explore different settings of the learning schemes by considering alternative choices of $\mathbf{p}_{j}$ in Eq.~\ref{chairp}, $\mu_{k}$ in Eq.~\ref{expertloss} and $\lambda$ in Eq.~\ref{finalloss}.
We summarize different variants in Table~\ref{tab:moe_variants}.

\jpei{In this work, we are focusing on context-to-act task~\cite{budzianowski2018multiwoz}, so natural language generation (NLG) baselines (e.g., SC-LSTM~\cite{wen2015semantically}) will not be taken into consideration.}

\vspace*{-0.5\baselineskip}
\subsection{Implementation details}
The vocabulary size is the same as in the original paper that releases the dataset \cite{budzianowski2018multiwoz}, which has 400 tokens.
Out-of-vocabulary words are replaced with ``$<$UNK$>$''.
We set the word embedding size to 50 and all \ac{LSTM} hidden state sizes to 150.
We use Adam~\cite{adam_optimizer} as our optimization algorithm with hyperparameters $\alpha=0.005$, $\beta1=0.9$, $\beta2=0.999$ and $\epsilon= 10^{-8}$. 
We also apply gradient clipping \cite{pmlr-v28-pascanu13} with range [--5, 5] during training.
We use $l2$ regularization to alleviate overfitting, the weight of which is set to 0.00001.
We set mini-batch size to 64.
We use greedy search to generate the response during testing. 
\jpei{Extra techniques (e.g., beam search) are not incorporated, because our main concern is the modular model outperforms single-module model instead of the effectiveness of these popular techniques.}

\vspace*{-0.5\baselineskip}
\subsection{Dataset}
\label{subsec:dataset}
Our experiments are conducted on the \acf{MultiWOZ}~\cite{budzianowski2018multiwoz} dataset. 
This is the latest large-scale human-to-human \ac{TDS} dataset with rich semantic labels (e.g., domains and dialogue actions) and benchmark results of response generation.\footnote{http://dialogue.mi.eng.cam.ac.uk/index.php/corpus/}
\ac{MultiWOZ} consists of $\sim$10k natural conversations between a tourist and a clerk. 
We consider 6 specific action-related domains (i.e., \textit{Attraction}, \textit{Hotel}, \textit{Restaurant}, \textit{Taxi}, \textit{Train}, and \textit{Booking}) and 1 universal domain (i.e., \textit{General}). 
67.37\% of dialogues are cross-domain which covers 2--5 domains on average. 
The average number of turns per dialogue is 13.68 and a turn contains 13.18 tokens on average.
To facilitate reproducibility of the results, the dataset is randomly split into into 8,438/1,000/1,000 dialogues for training, validation, and testing, respectively.

\vspace*{-0.5\baselineskip}
\subsection{Evaluation metrics}
We use three commonly used evaluation metrics \cite{budzianowski2018multiwoz}:
\begin{itemize}[leftmargin=*,align=left,nosep]
    \item \textit{Inform.} The fraction of responses that provide a correct entity out of all responses.
    \item \textit{Success.} The fraction of responses that answer all the requested attributes out of all responses.
    \item \textit{BLEU.} This is a score for comparing a generated response to one or more reference responses.
\end{itemize}
Following~\citet{budzianowski2018towards}, we use \textit{Score} = 0.5*\textit{Inform}+ 0.5*\textit{Success}+\textit{BLEU} as the selection criterion to choose the best model on the validation set and report the performance of the model on the test set. 
We utilize a paired t-test to show statistical significance ($p <0.01$) of relative improvements.
\section{Results}
\label{sec:results}

This section describes the results of our experiments and answers research questions proposed in \S\ref{sec:experimental-setup}.

\vspace*{-0.5\baselineskip}
\subsection{Performance of single-module \acp{TDS} on different domains (RQ1)}
\label{subsec:e3_bsl_domain}

To answer \textbf{RQ1}, we assess the performance of the single-module baseline \acs{S2SAttnLSTM} and its three variants with settings (V1, V2, and V3) described in \S\ref{subsec:bsls} on different domains.
The results are shown in Table~\ref{tab:bsls_on_domain}.

We can see that none of those models can consistently outperform the others on all domains and all metrics.
That is to say, a model can achieve its best performance only in some particular situations.
To be specific, \acs{S2SAttnLSTM} achieves its best performance only on the \textit{Hotel} domain in terms of \textit{BLEU} and \textit{Taxi} domain in terms of \textit{Success}. 
\acs{S2SAttnLSTM}/V1 outperforms all other models on the \textit{Restaurant} domain on all metrics and on the \textit{Hotel} domain (except for \textit{BLEU}).
\acs{S2SAttnLSTM}/V2 beats the others on the \textit{Attraction} and \textit{Taxi} domains in terms of all metrics.
\acs{S2SAttnLSTM}/V3 performs best on the \textit{Booking} and \textit{General} domains in terms of \textit{BLEU} and \textit{Score}.
Overall, \acs{S2SAttnLSTM}/V1 specializes in, and leads on, the \textit{Hotel} and \textit{Restaurant} domains.
\acs{S2SAttnLSTM}/V2 acts as an expert bot specialized for the \textit{Attraction}, \textit{Taxi}, \textit{UNK} domains, and \acs{S2SAttnLSTM}/V2 serves as an expert bot for the \textit{Train}, \textit{Booking}, \textit{General} domains.
Generally, the experimental results verify the assumption and motivation of our \ac{MTDS} framework.

\vspace*{-0.5\baselineskip}
\subsection{Overall performance (RQ2)}
\label{sec:e1_comparsion}

To answer our main research question, \textbf{RQ2}, we evaluate the performance of \ac{TokenMoE} and the baselines (\ac{S2SAttnLSTM}, \ac{TokenMoE} and their variants with settings V1, V2, V3).
The results are shown in Table~\ref{tab:moe_vs_bsl}.
\begin{table}[htb!]
\vspace*{-1\baselineskip}
\setlength{\tabcolsep}{2pt}
\caption{Comparison between \ac{TokenMoE}, the benchmark baseline \acs{S2SAttnLSTM}, and their variant models using setting V1, V2, V3, respectively. Bold results indicate a statistically significant improvement over the strongest baseline (paired t-test, $p < 0.01$).}
\label{tab:moe_vs_bsl}
\centering
\begin{tabular}{lllll}
\toprule
        & {\bfseries Inform (\%)} & {\bfseries Success (\%)} & {\bfseries BLEU (\%)}  & {\bfseries Score} \\
\midrule
\acs{S2SAttnLSTM}~\cite{budzianowski2018multiwoz} & 67.20       & 57.20        & 17.83 & 80.03 \\
\acs{S2SAttnLSTM}/V1   & 63.60 & 52.20 & 18.10 & 76.00 \\
\acs{S2SAttnLSTM}/V2   & 67.20 & 58.90 & \textbf{20.85} & 83.90 \\
\acs{S2SAttnLSTM}/V3   & 68.60 & 59.30 & 19.41 & 83.36 \\
\ac{TokenMoE}/V1   & 64.00 & 52.50 & 18.95 & 77.20 \\
\ac{TokenMoE}/V2   & 62.60 & 54.30 & 18.90 & 77.35 \\
\ac{TokenMoE}/V3   & 62.90 & 54.00 & 18.34 & 76.79 \\
\midrule
\ac{TokenMoE}      & \textbf{75.30} & \textbf{59.70} & 16.81 & \textbf{84.31} \\
\bottomrule
\end{tabular}
\vspace*{-1\baselineskip}
\end{table}



First, \ac{TokenMoE} outperforms all baseline models by a large margin in terms of all metrics. 
Especially, \ac{TokenMoE} significantly outperforms the benchmark single-module baseline \ac{S2SAttnLSTM}, by 8.1\% of \textit{Inform} and 2.5\% of \textit{Sucecess}, which maintains the same settings as the original paper~\cite{budzianowski2018multiwoz}. 
This shows that \ac{TokenMoE} has an advantage of task completion by providing more appropriate entities and answering the requested attributes as many as possible. 

Second, \ac{TokenMoE} greatly outperforms \ac{TokenMoE}/V1 by 11.7\% on \textit{Inform} and 7.5\% on \textit{Success}.
This is true with \ac{S2SAttnLSTM} and \ac{S2SAttnLSTM}/V1 except that the improvements are smaller, i.e., 3.6\% on \textit{Inform} and 5.0\% on \textit{Sucecess}.
On the one hand, this means that the attention mechanism is effective.
On the other hand, this also shows that the attention mechanisms under our \ac{TokenMoE} can be more effective and have an even more important role to play. 
That is to say, the \ac{MTDS} framework has more potential to improve by separating the modeling of expert and chair bots.
\ac{TokenMoE}/V2 is inferior to \acs{S2SAttnLSTM}/V2 when changing the \acs{LSTM} cell as \acs{GRU}.
Similarly, \ac{TokenMoE}/V3 is less effective than \acs{S2SAttnLSTM}/V3 when decreasing the number of hidden units.
This indicates that \ac{TokenMoE} is more sensitive to the number of parameters, which we think is due to the fact that \ac{TokenMoE} has a hard time learning the expert bots and their coordination with a small parameter space.

Third, all models achieve about 10\% higher values in terms of \textit{Inform} than in terms of \textit{Success}.
This shows that the big challenge of the dialogue generation task is how to answer requested attributes in a real-time manner. 
The \textit{BLEU} scores of all models are quite low compared with the state-of-the-art result (45.6\%) of machine translation~\cite{edunov2018understanding} but are similar to the state-of-the-art result (18.9\%) for dialogue generation~\cite{budzianowski2018multiwoz}.
This supports prior claims that the $BLEU$ score is not an ideal measurement for dialogue generation and explains the reason why we use $Score$ to choose our best model. 
\jpei{Table \ref{tab:case_study} shows an example of the baseline \acs{S2SAttnLSTM} and \acs{TokenMoE} output, which indicates that a lower $BLEU$ still does mean more appropriate response with more detail information.}
\begin{table}[htb!]
\setlength{\tabcolsep}{2pt}
\caption{An example of the generated responses of \acs{S2SAttnLSTM} and \acs{TokenMoE}. A user would prefer to get detail information of the train before booking a ticket.}
\label{tab:case_study}
\centering
\begin{tabular}{ll}
\toprule
Model       & Response \\ \midrule
S2SAttnLSTM & \begin{tabular}[c]{@{}l@{}}i have {[}value\_count{]} trains that match your criteria . \\ would you like me to book it for you ?\end{tabular} \\
TokenMOE    & \begin{tabular}[c]{@{}l@{}}i have train {[}train\_id{]} that leaves at {[}value\_time{]} and \\  arrives at {[}value\_time{]} . would you like me to book it ?\end{tabular} \\ 
\bottomrule
\end{tabular}
\vspace*{-1\baselineskip}
\end{table}

\vspace*{-0.5\baselineskip}
\subsection{Exploration of learning schemes for \ac{TokenMoE} (RQ3)}
\label{subsec:e3_learning_schemes}

To address \textbf{RQ3}, we explore how token-level \ac{MOE} and learning schemes used in \ac{TokenMoE} affect the performance. 
In Table~\ref{tab:moe_schmes}, we report the results of four variants of \ac{TokenMoE} (see Table~\ref{tab:moe_variants}). 
The detailed settings of each variant are as follows:
\begin{itemize}[leftmargin=*,align=left,nosep]
    \item S1 regards $\mu_k$ and $\lambda$ as learnable parameters while the others regard them as hyperparameters.
    \item S2 uses token-level \ac{MOE} but does not use the global-and-local learning scheme, i.e., $\lambda=0$ in Eq.~\ref{finalloss}.
    \item S3 does not use token-level \ac{MOE} and directly uses the prediction probability of the chair bot, i.e., $\mathbf{p}_j=\mathbf{p}_j^{k+1}$ in Eq.~\ref{chairp}, which actually degenerates into \ac{S2SAttnLSTM} with the proposed global-and-local learning.
    \item S4 uses both token-level \ac{MOE} and global-and-local learning.
\end{itemize}

\begin{table}[h]
\setlength{\tabcolsep}{4pt}
\caption{Comparison of \ac{TokenMoE} with different learning schemes (S1, S2, S3, S4) and the benchmark baseline \acs{S2SAttnLSTM}. Bold results indicate a statistically significant improvement over the strongest baseline (paired t-test, $p < 0.01$).}
\label{tab:moe_schmes}
\centering
\begin{tabular}{lllll}
\toprule
        &{\bfseries Inform (\%)} & {\bfseries Success (\%)} & {\bfseries BLEU}  & {\bfseries Score} \\
\midrule
\acs{S2SAttnLSTM}/V2   & 67.20 & 58.90 & \textbf{20.85} & 83.90 \\
\midrule
\acs{TokenMoE}/S1      & 66.20 & 54.90 & 19.11 & 79.66 \\
\acs{TokenMoE}/S2      & 66.50 & 56.90 & 19.48 &  81.18 \\
\acs{TokenMoE}/S3      & 70.60 & \textbf{60.60} & 18.67 & 84.27 \\
\acs{TokenMoE}/S4      & \textbf{75.30} & 59.70 & 16.81 & \textbf{84.31}\\
\bottomrule
\end{tabular}
\vspace*{-0.8\baselineskip}
\end{table}

First, S1 is worse than the other three variants on all metrics, which shows that it is not effective to learn $\mu_k$ and $\lambda$.
The reason is that 
\acs{TokenMoE} may fall into the \textit{optimization trap} due to learning $\mu_k$ and $\lambda$. 
That is, \acs{TokenMoE} learns a very small weight for the local loss of each expert (i.e., $\mu_k \approx 0$) and a large weight for the global loss of the chair bot (i.e., $\lambda \approx 1$). Afterwards, this loss will never decrease any more, so the model learns nothing useful.

Second, S2 is even worse than \acs{S2SAttnLSTM}/V2 on all metrics which means the performance cannot be improved with the proposed token-level \ac{MOE} alone.
We believe the reason is that token-level \ac{MOE} makes the model harder to learn, i.e., the model needs to learn not only each prediction distribution by the expert and chair bots but also their combinations.
This can be verified by the fact that with token-level \ac{MOE} and global-and-local learning, S4 further improves \textit{Inform} by 4.7\% compared with S3.
Our explanation is that the global-and-local learning makes token-level \ac{MOE} easier to learn by incorporating supervisions on both the prediction distribution of each expert (local loss in Eq.~\ref{expertloss}) and their combination (global loss in Eq.~\ref{finalloss}).

Third, S3 is better than \acs{S2SAttnLSTM}/V2 in terms of \textit{Inform} and \textit{Success}.
Also, it achieves the best performance on \textit{Success}.
This shows that \acs{TokenMoE} with an appropriate scheme is expert in task accomplishment for a \acs{TDS}, i.e., \acs{TokenMoE}/S3 is able to generate more correct entities and answer more requested attributes.
The reason behind this is quite clear: with global-and-local learning, each expert is trained to specialize on a particular domain, which means the chair and the experts are able to extract more manifold candidate tokens, each of them holds a unique preference distribution over the output vocabulary. 
For example, a \textit{Booking} expert has a high probability to produce the intent-oriented token  ``booked'' in the response ``Your order has been booked''.  
In contrast, without global-and-local learning, the single model prefers to generate more generic tokens (e.g., ``thanks'') that occur most frequently in all domains.

However, it is worth noting that both S2 and S3 are worse than \acs{S2SAttnLSTM}/V2 on \textit{BLEU} and S4 is even worse than S2 and S3.
This indicates that token-level \ac{MOE} and global-and-local learning have a negative influence on the response fluency evaluated by \textit{BLEU}.
A possible reason is that various candidate tokens from the chair and experts make the dialogue contexts more complex, which increases the difficulty of generating a fluent response.
Another reason is that \textit{BLEU} is not an ideal metric for dialogue generation task, as we discussed in \S\ref{sec:e1_comparsion}.

\section{Related Work}

There are two dominant frameworks for \acp{TDS}: modularized pipeline \acp{TDS} and end-to-end single-module \acp{TDS}. 

\vspace*{-0.5\baselineskip}
\subsection{Modularized pipeline \acp{TDS}}
Modularized pipeline \acp{TDS} frameworks consists of a pipeline with several modules.
Examples include \ac{NLU}~\cite{chen2017nlu,bapna2017pc_slu}, \ac{DST}~\cite{zhong2018dst,rastogi2018pc_dst}, \ac{PL}, and \ac{NLG}~\cite{duvsek2016nlg,yi2019pc_dg}. 
Each module has an explicitly decomposed function for a specialized subtask, which is beneficial to track errors.
\citet{young2013pomdp} summarize typical pipeline \acp{TDS} that are constitutive of distinct modules following a \acs{POMDP} paradigm.
\citet{crook2016tds_platform} develop a \ac{TDS} platform that is loosely decomposed into three modules, i.e., initial processing of input, dialogue state updates, and policy execution.
\citet{yan2017building} present a \ac{TDS} for completing various purchase-related tasks by optimizing individual upstream-dependent modules, i.e., query understanding, state tracking and dialogue management.
However, the pipeline setting of these methods will unavoidably incur upstream propagation problem~\cite{chen2017survey}, module interdependence problem~\cite{chen2017survey} and joint evaluation problem~\cite{young2013pomdp}. 
Unlike the methods listed above, our \acs{MTDS} constists of a group of modules including a chair bot and several expert bots.
This design addresses the module interdependence problem since each module is independent among the others. 
Besides, the chair bot alleviates the error propagation problem because it is able to manage the overall errors through an effective learning schemes.

\vspace*{-0.5\baselineskip}
\subsection{End-to-end single-module \acp{TDS}}
End-to-end single-module systems address the \acs{TDS} task with only one module, which maps a \textit{dialogue context} to a \textit{response} directly \cite{wen2017network}.
There is a growing focus in research on end-to-end approaches for \acp{TDS}, which can enjoy global optimization and facilitate easier adaptation to new domains~\cite{chen2017survey}.
\citet{sordoni2015neural} show that using an \ac{RNN} to generate text conditioned on the dialogue history results in more natural conversations. 
Later improvements have been made by adding an attention mechanism~\cite{vinyals2015neural,li2016persona}, by modeling the hierarchical structure of dialogues~\cite{serban2016building}, or by jointly learning belief spans~\cite{lei2018sequicity}. 
However, existing studies on end-to-end \acp{TDS} mostly use a single-module underlying model to generate responses for complex dialogue contexts.
This is practically problematic because dialogue contexts are very complicated with multiple sources of information~\cite{chen2017nlu}. 
In addition, previous studies show that it is abnormal to find a single model that achieves the best results on the overall task based on empirical studies from different machine learning applications~\cite{dietterich2000ensemble,masoudnia2014mixture}.

Different from the methods listed above, which use a single module to achieve \acp{TDS}, our \acs{MTDS} uses multiple modules (expert and chair bots), which makes good use of the specialization of different experts and the generalization of chair for combining the final outputs.
Besides, our \acs{MTDS} model is able to track who is to blame when the model makes a mistake.

\section{Conclusion and Future Work}
\label{sec:conclusion}

This paper we have presented a neural \acf{MTDS} framework composed of a chair bot and several expert bots. 
We have developed a \ac{TokenMoE} model under this \ac{MTDS} framework, where the expert bots make multiple token-level predictions at each timestamp and the chair bot predicts the final generated token by fully considering the whole outputs of all expert bots.
Both the chair bot and the expert bots are jointly trained in an end-to-end fashion.

We have conducted extensive experiments on the benchmark dataset \ac{MultiWOZ} and evaluated the performance in terms of four automatic metrics (i.e., \textit{Inform}, \textit{Success}, \textit{BLEU}, and \textit{Score}). 
We find that no general single-module \ac{TDS} model can constantly outperform the others on all metrics.
This empirical observation facilitates the design of a new framework, i.e., \ac{MTDS} framework.
We also verify the effectiveness of \ac{TokenMoE} model compared with the baseline using a single-module model.
Our \ac{TokenMoE} outperforms the best single-module model (\ac{S2SAttnLSTM}/V2) by 8.1\% of \textit{inform rate} and 0.8\% of \textit{success rate}.
Besides, it significantly beats \ac{S2SAttnLSTM}, the benchmark single-module baseline, by 3.5\% of \textit{Inform} and 4.2\% of \textit{Sucecess}.
In addition, the experimental results show that learning scheme is an important factor of our \ac{TokenMoE} model.

In the future work, we hope to explore \ac{SentenceMoE} and combine it with the current \ac{TokenMoE} to see whether the hybrid model will further improve the performance.
Besides, we plan to try more fine-grained expert bots (e.g., according to user intents or system actions) and more datasets to test our new framework and model.


\begin{acks}
This research was partially supported by
Ahold Delhaize,
the Association of Universities in the Netherlands (VSNU),
the China Scholarship Council (CSC),
and
the Innovation Center for Artificial Intelligence (ICAI).
All content represents the opinion of the authors, which is not necessarily shared or endorsed by their respective employers and/or sponsors.
\end{acks}

\bibliographystyle{ACM-Reference-Format}
\bibliography{WCIS2019-jiahuan-MTDS}

\end{document}